\documentclass[runningheads]{llncs}

\usepackage{eccv}

\usepackage{eccvabbrv}
\usepackage{stfloats}
\usepackage{graphicx}
\usepackage{booktabs}
\usepackage{tabularx}
\usepackage{cuted}
\usepackage{caption}
\usepackage{float}
\usepackage{makecell}
\usepackage{siunitx}
\usepackage{amssymb}
\usepackage{multirow}
\usepackage{array}
\usepackage{algorithm}
\usepackage{algpseudocode}
\usepackage[accsupp]{axessibility}
\usepackage{orcidlink}

\usepackage{hyperref}

\makeatletter
\newcommand{\equalcontrib}{\textsuperscript{\normalfont\@fnsymbol{1}}}
\makeatother

\begin{document}

\title{Pistachio:\\Towards Synthetic, Balanced, and Long-Form Video Anomaly Benchmarks}

\titlerunning{Pistachio: A Synthetic Video Anomaly Benchmark}

\author{
Jie Li\inst{1,2}\thanks{Equal contribution. Work done during Jie Li's internship at SJTU.\\
\makebox[0pt][r]{\textsuperscript{$\dagger$}\ }Corresponding author.}\orcidlink{0009-0003-2055-7674} \and
Hongyi Cai\inst{3}\equalcontrib\orcidlink{0009-0002-9423-0749} \and
Mingkang Dong\inst{3}\orcidlink{0009-0000-0276-6075} \and
Muxin Pu\inst{4}\orcidlink{0009-0000-8597-1938} \and
\\
Shan You\inst{5} \and
Fei Wang\inst{5} \and
Tao Huang\inst{1}\textsuperscript{$\dagger$}\orcidlink{0000-0002-4463-4078}
}

\authorrunning{J.~Li et al.}

\institute{
Shanghai Jiao Tong University \and
University of Science \& Technology Beijing \and
Universiti Malaya \and
Monash University \and
SenseTime Research \\
\email{lj2085727892@163.com,huntocn@gmail.com}\\
}

\maketitle

\begin{center}
    \captionsetup{type=figure}
    \includegraphics[width=1.0\linewidth]{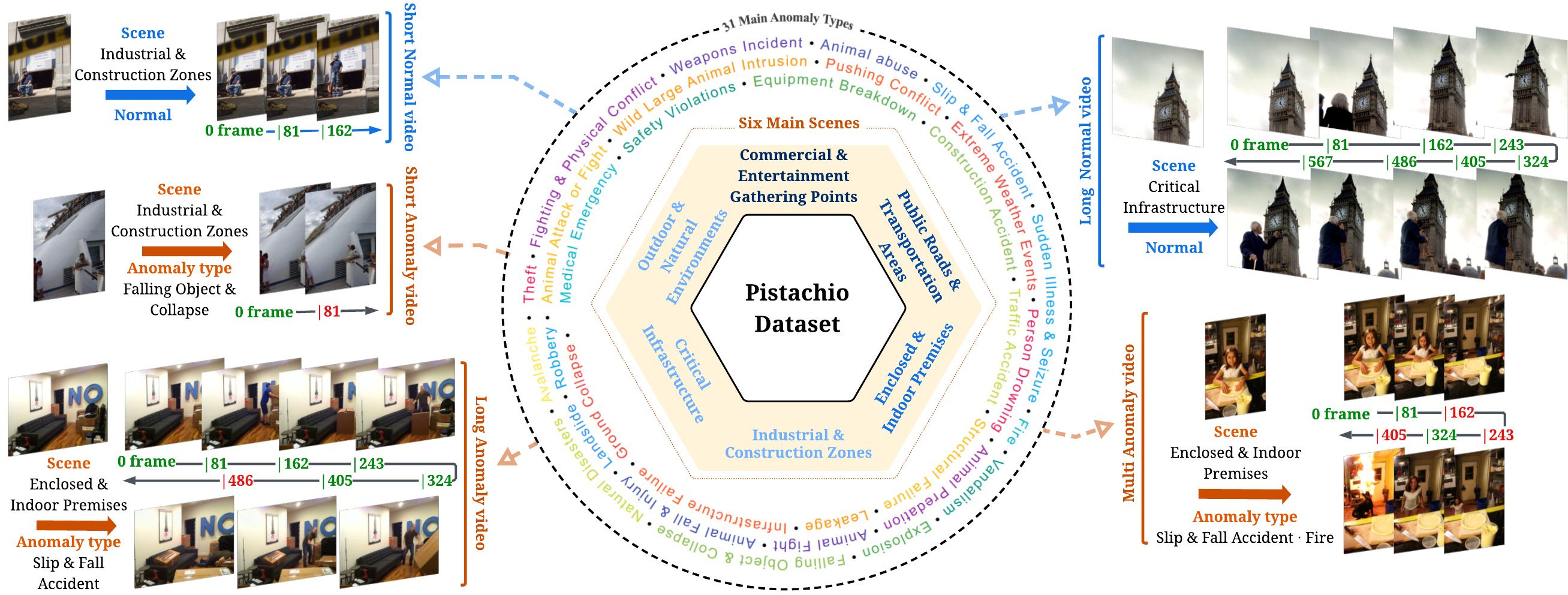}
    \captionof{figure}{\textbf{We introduce Pistachio -} a benchmark for video anomaly analysis, which aims at \textbf{two fundamental tasks}:
    VAD and VAU. The VAD dataset totals 1.6 million frames and extends existing datasets by expanding the number of scenes from hundreds to thousands, covering 31 distinct anomaly types, over half of which are unique to this benchmark. Pistachio offers multi-granularity annotations at both the event and video levels. The entire benchmark was produced via a highly automated pipeline.}
    \label{fig:1}
\end{center}

\begin{abstract}
  Automatically detecting abnormal events in videos is crucial for modern autonomous systems, yet existing Video Anomaly Detection (VAD) benchmarks lack the scene diversity, balanced anomaly coverage, and temporal complexity needed to reliably assess real-world performance. Meanwhile, the community is increasingly moving toward Video Anomaly Understanding (VAU), which requires deeper semantic and causal reasoning but remains difficult to benchmark due to the heavy manual annotation effort it demands. In this paper, we introduce Pistachio, a new VAD/VAU benchmark constructed entirely through a controlled, generation-based pipeline. By leveraging recent advances in video generation models, Pistachio provides precise control over scenes, anomaly types, and temporal narratives, effectively eliminating the biases and limitations of Internet-collected datasets. Our pipeline integrates scene-conditioned anomaly assignment, multi-step storyline generation, and a temporally consistent long-form synthesis strategy that produces coherent 41-second videos with minimal human intervention. Extensive experiments demonstrate the scale, diversity, and complexity of Pistachio, revealing new challenges for existing methods and motivating future research on dynamic and multi-event anomaly understanding. All open-source project assets are now publicly available at \url{https://pistachio-video.github.io}.
\end{abstract}

\section{Introduction}
\label{sec:intro}

With the increasing deployment of autonomous driving, drone inspection, and large-scale surveillance systems, the ability to automatically detect abnormal events in videos has become indispensable.VAD
\cite{tian2021weakly,Micorek_2024_CVPR,chen2022mgfn,URDMU_zh} aims to identify deviations from expected patterns, ranging from fires and traffic collisions to hazardous human activities. However, despite the steady growth of publicly available VAD datasets, existing benchmarks \cite{Acsintoae_CVPR_2022,zhang2016single,wang2021robust,ramachandra2020street} still fail to effectively and comprehensively evaluate modern VAD systems. Their limited diversity, biased anomaly distribution, and insufficient temporal complexity restrict both the reliability of evaluation and the generalization ability of trained models.

A closer examination reveals that these issues largely stem from how current datasets are sourced and curated. (i) Most videos are collected from the Internet, resulting in a heavy bias toward urban roadways and public streets. (ii) Anomaly types reflect events that most easily gain online visibility, producing severe long-tail distributions—frequent violent incidents but scarce environmental hazards, infrastructure failures, or rare accidents. (iii) The ``normal'' portions of the datasets are overly simplistic, typically featuring walking or light movement, which fails to capture the diversity of ordinary human or environmental activities. These weaknesses collectively cause VAD models to overfit to dataset-specific patterns, confuse complex-but-normal behaviors with anomalies, and fail to recognize rare or subtle events.

Beyond detection, the community has recently shifted toward VAU ~\cite{tang2024hawklearningunderstandopenworld,du2024uncoveringwhathowcomprehensive,zhang2025holmesvaulongtermvideoanomaly}, which emphasizes deeper semantic and causal reasoning of identifying what occurred, why it occurred, and how the event unfolds over time. Yet current VAD benchmarks are intrinsically unable to evaluate these capabilities, because they lack structured temporal narratives, multi-step event progressions, and causal dependencies.
Moreover, constructing a VAU benchmark is prohibitively expensive: it requires dense event-level annotations, temporal segmentation, and detailed semantic descriptions, especially when multiple anomalies co-occur within a single video. 
As a result, existing benchmarks remain limited in scope, failing to provide a comprehensive or scalable evaluation framework that fully captures the complexity of VAU.

In this paper, we propose \textit{Pistachio}, a new VAD/VAU benchmark constructed entirely through a controlled, generation-based pipeline. Motivated by the remarkable progress of modern video generation models such as Sora \cite{liu2024sora}, Veo 3 \cite{wiedemer2025videomodelszeroshotlearners}, and Wan \cite{wan2025wanopenadvancedlargescale}, we argue that synthetic generation offers a more deterministic and scalable solution than filtering long-tailed Internet videos and performing labor-intensive manual annotation.
Generation-based construction allows precise control over scenes, anomaly types, temporal progression, and visual diversity, providing balanced anomaly coverage and eliminating dataset biases that plague existing VAD resources.

Specifically, we propose several crucial designs to address the challenges in building a reliable, long-form anomaly benchmark. First, we develop a scene-conditioned anomaly assignment strategy that leverages Vision-Language Models (VLMs) to categorize source images into six major scene types and allocate contextually appropriate anomaly types, ensuring realistic and logically consistent pairings.
Second, we design a multi-step storyline generation framework that decomposes each video into 7–8 descriptive segments for long videos (2–3 for short videos), enabling coherent narrative progression from normal activities to anomalous events.
Third, we introduce a temporally consistent long-form video synthesis mechanism that chains short video clips using the last frame of each segment as the starting frame for the next, maintaining visual continuity across 41-second long videos (16-second for short videos).
Finally, we implement a hybrid human-AI filtering approach that combines automated quality assessment with expert verification to ensure generated videos are free from artifacts and conform to real-world logic.

The resulting Pistachio-VAD benchmark is, to our knowledge, the most category-rich anomaly dataset to date, containing 4,962 videos spanning 1.68M frames, across 6 major scene categories and 31 diverse anomaly types, many of which do not appear in any prior work (e.g., landslides, animal predation, equipment breakdown). The dataset includes both static and moving cameras and features complex normal behaviors absent in existing benchmarks.
In parallel, Pistachio-VAU contains 1,385 videos with event-level and video-level descriptions, including 35 videos with multiple co-occurring anomalies, which provides the first large-scale, fully annotated VAU benchmark constructed with zero manual annotation.

Our contributions can be summarized in three folds:
\begin{itemize}
    \item We introduce Pistachio-VAD, a scalable, generation-based VAD benchmark that breaks scene and anomaly biases. It provides diverse scenes, balanced anomaly categories, and complex normal behaviors in large-scale, long-form videos.
    \item We design a fully automated pipeline for creating high-quality long-form anomaly videos. The controlled generation framework produces coherent 41-second videos through Scene-Aware Classification, Anomaly Type Specification, Multi-step storyline generation, and Temporally consistent long-form video synthesis, requiring minimal human intervention.
    \item We construct Pistachio-VAU, the first large-scale VAU benchmark without manual annotation. We use generation storylines as structured semantic annotations, enabling comprehensive anomaly understanding including multi-event reasoning, without any manual labeling effort.
\end{itemize}

\begin{table*}[t]
\centering
\setlength{\tabcolsep}{6pt} 
\renewcommand{\arraystretch}{1.0} 
\caption{Overview of existing VAD datasets.}
\label{tab:dataset_overview}
\resizebox{\textwidth}{!}{ 
\renewcommand{\theadfont}{\normalsize\bfseries}
\small 
\begin{tabular}{l c S[table-format=4.0] S[table-format=8.0] S[table-format=4.0] l S[table-format=2.0] c}
\toprule

\textbf{Dataset} & \textbf{Year} & \multicolumn{1}{c}{\textbf{Video}} & \multicolumn{1}{c}{\textbf{Frames}} & \multicolumn{1}{c}{\textbf{Scenes}} & \textbf{Annotation} & \multicolumn{1}{c}{\textbf{\makecell{Detailed \\ Anomaly types}}} & \textbf{Scalability} \\
\midrule
Subway Exit \cite{adam2008robust} & 2008 & 1 & 38940 & 1 & Frame & 3 & $\times$ \\
Subway Entrance \cite{adam2008robust} & 2008 & 1 & 86535 & 1 & Frame & 5 & $\times$ \\
UCSD Ped1 \cite{li2013anomaly} & 2010 & 70 & 14000 & 1 & Frame & 5 & $\times$ \\
UCSD Ped2 \cite{li2013anomaly} & 2010 & 28 & 4560 & 1 & Frame & 5 & $\times$ \\
CUHK Avenue \cite{abnormal2013lu} & 2013 & 35 & 30652 & 1 & Frame & 5 & $\times$ \\
ShanghaiTech \cite{luo2017revisit} & 2017 & 437 & 317398 & 11 & Frame & 11 & $\times$ \\
UCF-Crimes \cite{sultani2019realworldanomalydetectionsurveillance} & 2018 & 1900 & 13741393 & {N/A} & Frame & 12 & $\times$ \\
XD-Violence \cite{wu2020not} & 2020 & 4754 & 114096 & {N/A} & Frame & 5 & $\times$ \\
Street Scene \cite{ramachandra2020street} & 2020 & 81 &  203257 & 17 & Frame & 5 & $\times$ \\
UBnormal \cite{acsintoae2022ubnormal} & 2022 & 543 & 236902 & 22 & Pixel+Frame & 22 & $\times$ \\
NWPU Campus \cite{cao2023new} & 2023 & 547 & 1466073 & 43 & Frame & 28 & $\times$ \\
MSAD \cite{zhu2024advancing} & 2024 & 720 & 447236 & 14 & Frame & 11 & $\times$ \\
Pistachio (Ours) & 2026 & \textbf{4962} & 1676822 & \bfseries \textbf{3896} & \textbf{Frame+text} & \textbf{31} & $\checkmark$ \\
\bottomrule
\end{tabular}
} 
\end{table*}

\section{Related work}
\label{sec:formatting}

\subsection{Video Anomaly Detection}

\textbf{VAD methods.} Video Anomaly Detection methods~\cite{wu2024openvocabularyvideoanomalydetection,yang2024followrulesreasoningvideo,zanella2024harnessinglargelanguagemodels,semi1,semi2,full1,full2,weak1,weak2,weak3,weak4,weak5} can be categorized into semi-supervised \cite{semi1,semi2}, weakly-supervised \cite{weak1,weak2,weak3,weak4,weak5}, and fully supervised \cite{full1,full2} methods. Recently, VLM- and LLM-based VAD has gained momentum. OVVAD~\cite{wu2024openvocabularyvideoanomalydetection} synthesizes pseudo-unseen anomaly samples using a Novel Anomaly Synthesis module built on large generative vision models. Rule-based reasoning~\cite{yang2024followrulesreasoningvideo} converts normal frames into textual descriptions and uses LLMs to derive rules for identifying anomalies. Training-free pipelines~\cite{zanella2024harnessinglargelanguagemodels} combine VLMs and LLMs for inference without model training. As VAD evolves toward richer contextual and causal reasoning, these text-driven approaches highlight the importance of semantic annotations: an aspect our benchmark directly supports.

\textbf{VAD datasets.} 
We report several statistics about the most utilized datasets in Table \ref{tab:dataset_overview}. As these statistics reveal, despite extensive research, existing datasets remain insufficient for evaluating real-world anomaly detection.
Early benchmarks like CUHK Avenue \cite{abnormal2013lu}, UCSD Ped2 \cite{li2013anomaly}, and Street Scene \cite{ramachandra2020street} focus on single-scene environments and simple human activities, limiting generalization. Even with more classes, Street Scene \cite{ramachandra2020street} suffers from severe long-tail imbalance (e.g., 61 jaywalking vs. 1 motorcycle-on-sidewalk). ShanghaiTech \cite{luo2017revisit} expands the scene types but still includes only 158 anomaly instances across 11 categories. Larger datasets like UCF-Crime~\cite{sultani2019realworldanomalydetectionsurveillance} and MASD \cite{zhu2024advancing} incorporate more realistic anomalies, yet their Internet- and surveillance-sourced videos inevitably reflect scene bias and overrepresented event types, failing to capture subtle or rare anomalies. NWPU \cite{cao2023new} Campus attempts to alleviate these issues through staged events performed by volunteers, but this approach is labor-intensive and still cannot cover the full diversity of abnormal behaviors. Overall, the scarcity and long-tail nature of real abnormal events make dataset construction inherently challenging. Our generation-based approach provides a scalable and controllable alternative, enabling balanced anomaly coverage, diverse scenes, and complex normal behaviors that are difficult to obtain from Internet-collected data.

\subsection{Video Generation Models}
Recent advances in text-to-video generation \cite{hunyuan,wan2025wanopenadvancedlargescale,liu2024sora,wiedemer2025videomodelszeroshotlearners} have dramatically improved the realism, controllability, and temporal coherence of synthetic videos. Diffusion- \cite{peebles2023scalable} and diffusion-transformer-based \cite{peebles2023scalablediffusionmodelstransformers} models now support high-resolution, instruction-following video generation from text or images, often with strong physical plausibility. For example, OpenAI’s Sora \cite{liu2024sora} can generate minute-scale, photorealistic videos directly from prompts and input images, maintaining complex camera motion and scene dynamics. Google’s Veo series \cite{wiedemer2025videomodelszeroshotlearners} similarly produces high-quality clips with improved modeling of real-world physics and human motion, targeting cinematic applications. In parallel, open and industrial models such as Wan \cite{wan2025wanopenadvancedlargescale} adopt diffusion transformer architectures and large-scale pre-training to deliver efficient 720p, 24fps video generation for both text-to-video and image-to-video settings. These systems demonstrate strong visual fidelity over short durations (typically 5–10 seconds), but naively extending them to long videos often leads to temporal drift, artifacts, or prompt forgetting. Our work builds on this line of research by introducing a long-form generation mechanism that composes multiple short clips into coherent 41-second narratives.

\section{Pistachio Benchmark}
\label{sec:method}
\begin{figure*}[h!]
    \centering
    \includegraphics[width=\textwidth]{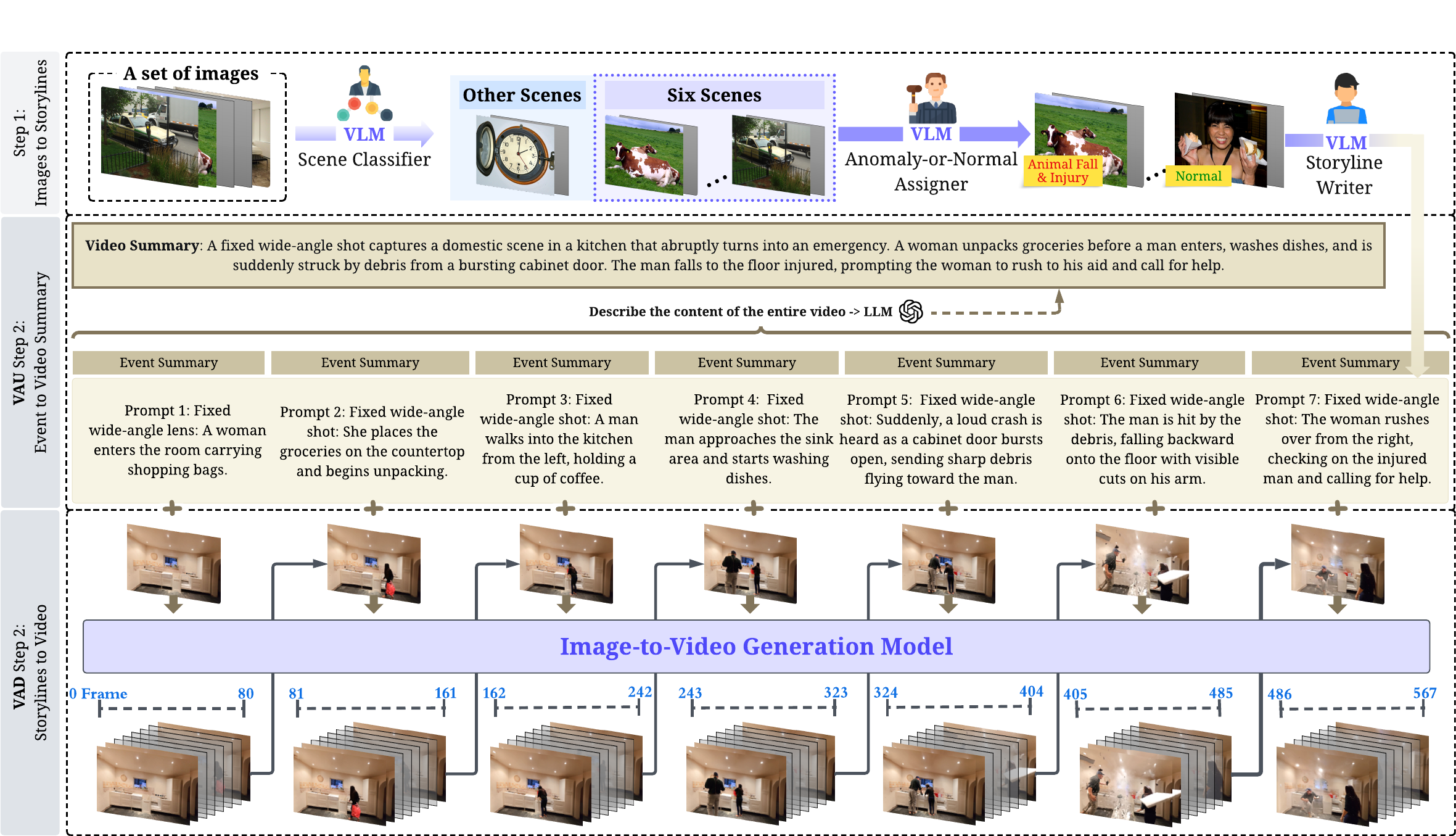}
    \caption{Overview of our video anomaly dataset generation pipeline. \textbf{Step 1 (Storyline Generation):} A VLM-based scene classifier processes input images to identify scenes. The Anomaly Type Allocator then assigns appropriate anomaly types (e.g., Fire) to generate coherent storylines across six scenes by LLMs. \textbf{VAU Step 2 (Event-to-Video Summary):} The storyline is decomposed into event summaries, with each event described by detailed prompts specifying camera angles, actions, and temporal progression. \textbf{VAD Step 2 (Storyline-to-Video):} An image-to-video generation model synthesizes coherent video sequences from the event summaries, producing temporally consistent frames that maintain narrative continuity across the entire storyline.}
    
    \label{fig:my-label}
\end{figure*}
\subsection{Dataset Statistics}
\subsubsection{Pistachio for Video Anomaly Detection}
We introduce our new dataset for VAD, a comprehensive collection comprising 4,962 videos, which total 29.11 hours of footage captured at 16 FPS. 
It is the largest known semi-supervised VAD dataset.
The dataset is structured across 6 major scenes and encompasses 31 distinct anomaly categories. 
Notably, we introduce 10 novel anomaly types that are absent in existing VAD benchmarks: Animal Predation, Animal Abuse, Construction Accidents, Infrastructure Failure, Natural Disasters, Avalanche, Equipment Breakdown, Extreme Weather Events, Ground Collapse, and Landslide.
In addition to frame-level labels, it provides rich textual annotations.
The detailed composition of our training and testing sets is provided in the supplementary material.

\begin{figure*}[h!]
    \centering
    \includegraphics[width=1.0\textwidth]{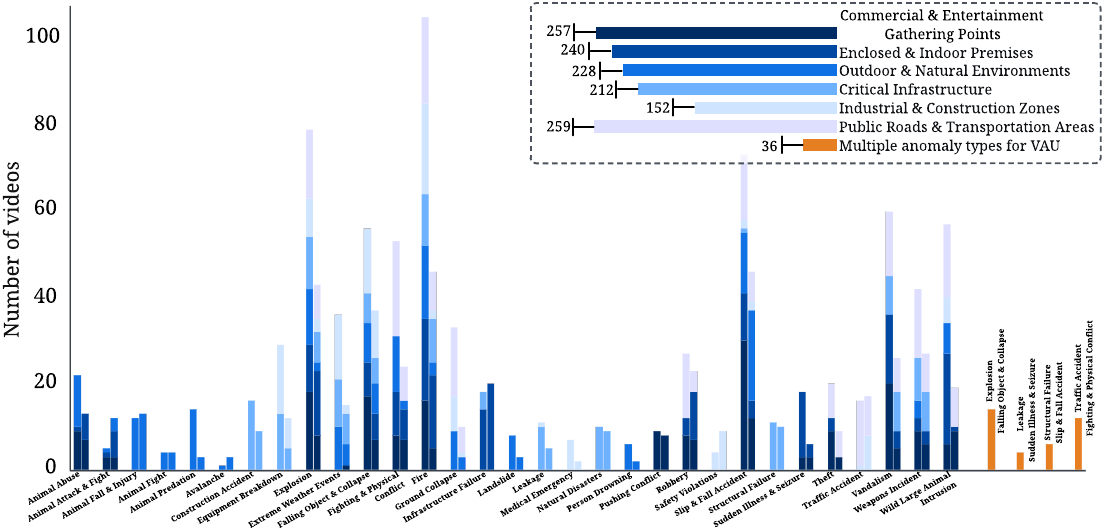}
    \caption{Distribution of anomaly videos across different anomaly types. For each type, the left bar represents short videos and the right bar represents long videos. The rightmost single bars indicate multi-anomaly videos, which are exclusively long-form.}
    \label{fig:upset_diagram}
\end{figure*}

To highlight the contributions of our dataset, we comprehensively compare it to existing benchmarks. 
Our dataset is distinguished by several outstanding traits. 
First, as illustrated in \ref{fig:upset_diagram}, it introduces a wealth of scene-specific anomalies, like Avalanche, Landslide, and Animal Predation. And our dataset integrates novel categories—nearly half of which are absent in other datasets—to emphasize critical social and ecological safety issues. We have maintained a balanced categorical distribution, directly addressing the persistent long-tail problem.
Second, the dataset's unprecedented scale in scene diversity comprehensively mitigates the long-standing issue of model scene-dependency. 
Third, as shown in the examples in Fig \ref{fig:1} our normal samples extend beyond trivial actions; thanks to a meticulous storyline design, we incorporate a rich array of complex daily activities, like handshaking and vehicle boarding. 
Fourth, 
Finally, the inclusion of both conventional static shots to mimic surveillance perspectives and dynamic mobile camera shots makes our dataset uniquely suited for emerging applications in fields such as drone monitoring and autonomous driving.

\subsubsection{Pistachio for Video Anomaly Understanding}

For the task of Video Anomaly Understanding, we have curated a specialized dataset comprising 1,385 videos, which total 517,514 frames. A key innovation of this dataset is its introduction of multiple, distinct anomaly types within a single video—a first for VAU benchmarks. Furthermore, each video is supported by comprehensive event-level and video-level annotations. Echoing the strengths of its VAD counterpart, our VAU dataset also stands as the most extensive benchmark to date in terms of scene richness, anomaly diversity, and balanced categorical distribution.

\subsection{Multi-Stage Data Generation Pipeline}
\label{sec:pipeline}
Our multi-stage data generation pipeline is meticulously designed to create a large-scale, high-quality, and richly annotated dataset.
The entire process is divided into three key stages. 

First, Video Data Generation Sec \ref{sec:video_data_generation} details the core synthesis pipeline. As shown in Fig.~\ref{fig:my-label}, this process involves three key steps: Step 1 uses a VLM to generate scene-conditioned anomaly storylines from static images, Step 2 decomposes these storylines into detailed event-level prompts, and Step 3 employs a video generation model to synthesize these into the final video clips.
This is followed by Video Data Filtering Sec \ref{sec:video_filtering} , a crucial quality control step that employs a hybrid human-AI screening approach to ensure the realism and logical consistency of all generated videos. Finally, and specifically to support the VAU task, the Annotation Generation stage Sec \ref{sec:annotation_generation} leverages an LLM/VLM to automatically process the storylines, generating the rich, multi-granularity (event-level and video-level) annotations.

\subsubsection{Video Data Generation}

\label{sec:video_data_generation}
\paragraph{Scene-Aware Classification.}
\label{sec:scene_aware_classification}
We utilize images $\mathcal{I} = \{I_1, \cdots, I_N\}$ sourced from the COCO 2017 dataset. A Vision-Language Model $\mathcal{M}(\cdot)$ (VLM) categorizes each image $I_i$ into one of $K=6$ predefined scene categories extracted from scene configuration $\mathcal{C}_{scene}$, plus an additional ``Other'' category, yielding $C = \{c_1, \cdots, c_K\} \cup \{\text{``Other''}\}$. The categorization prompt is provided in the supplementary materials.

This stage assigns each image to a scene group:
\begin{equation}
I_i \to \mathcal{G}_{\hat{c}_i}, \text{ where } \hat{c}_i = \mathcal{M}(I_i, \text{prompt}_{\text{scene}})
\end{equation}
where $\mathcal{G}_{c_j}$ denotes the set of images assigned to scene category $c_j$.

\paragraph{Anomaly Type Specification.}
\label{sec:anomaly_type}
For each scene category $c_j \in C$, we manually define a tailored set of plausible anomaly types $A_j = \{a_{j,1}, \cdots, a_{j,M_j}\}$ in $\mathcal{C}_{scene}[c_j]$ (detailed in supplementary materials). The VLM assigns specific anomaly types to each image within the scene group through an anomaly mapping $\mathcal{A}: I_i \mapsto a_i$:
\begin{equation}
\hat{a}_i = \mathcal{M}(I_i, \phi_j), \text{ where } \phi_j = \text{BuildPrompt}(c_j, A_j)
\end{equation}
The assignment prompt templates are provided in the supplementary materials.

\paragraph{Multi-step storyline generation.}
\label{sec:multi_step_storyline}
For each image $I_i$ with assigned scene $c_i$ and anomaly type $a_i$, we retrieve the corresponding prompt template $p_i = \mathcal{C}_{scene}[c_i][a_i]$ and format it to generate a coherent video storyline:
\begin{equation}
\mathcal{S}_i = \mathcal{M}(I_i, \psi_i), \text{ where } \psi_i = \text{FormatPrompt}(p_i, I_i)
\end{equation}
Each storyline $\mathcal{S}_i = \{s_1, \cdots, s_L\}$ comprises $L$ descriptive segments, where $L \in [7,8]$ for long videos and $L \in [2,3]$ for short videos. Detailed prompt templates are in the supplementary materials.

\paragraph{Temporally consistent long-form video synthesis.}
The final stage synthesizes video clips from the generated storylines. For each storyline $\mathcal{S}_i$, its segments $\{s_1, s_2, \dots, s_L\}$ are sequentially fed into the Wan video generation model. For each segment $s_l$, a video clip $V_l = \text{Wan}(I_{l-1}, s_l)$ is generated, where $I_{l-1}$ is the starting frame. The initial frame $I_0$ is the original image $I_i$ from $\mathcal{I}$, and for subsequent segments ($l > 1$), the last frame of $V_{l-1}$ serves as $I_{l-1}$ to ensure temporal continuity. All clips $\{V_1, V_2, \dots, V_L\}$ are concatenated to form the final video $V_{\text{final}}^i$.

\subsubsection{Video Data Filtering}
\label{sec:video_filtering}

 Balanced and highly realistic AI-generated data is essential for video anomaly detection. Our video filtering process consists of two stages: first, we employ VideoScore~\cite{he2024videoscore} to automatically score all generated videos across multiple quality dimensions, discarding those in the bottom 5\% (average score below 3.98) as a preliminary quality control step. Subsequently, we conduct rigorous manual filtering to ensure the remaining videos meet the following criteria:
\begin{enumerate}
    \item The anomaly events in the videos are reasonably designed and conform to real-life logic.
    \item No obvious signs of AI generation, such as object inconsistencies or sudden visual distortions.
    \item The camera angles, viewpoints, and lighting conditions are realistic and emulate those of genuine videos.
\end{enumerate}

This rigorous two-stage filtering ensures high visual fidelity and physical consistency across the dataset, guaranteeing that the final videos are free from generative artifacts and suitable for robust anomaly analysis.

\subsubsection{Annotation Generation and Refinement}
\label{sec:annotation_generation}

To ensure maximum rigor and accuracy, we adopt a hybrid annotation strategy. For VAD task, we utilize manual human annotation to define ground truth labels. For VAU task, we implement a multi-stage pipeline to generate event-level and video-level text annotations.

Our annotation workflow originates from the initial video storylines, which provide inherently precise, localized event-level descriptions. 
We utilize these texts both as the input prompts for video generation and as the foundational ground truth for anomaly understanding.
To generate holistic video-level descriptions, we utilize LLMs to summarize individual event annotations into a single, coherent video summary, as illustrated in Fig. \ref{fig:my-label}, Step 2.
This process accounts for temporal partitions and contextual variances within the storyline.

To further enhance the fidelity of the generated text, we introduce a VLM Refiner. 
The VLM performs a secondary pass, cross-referencing the generated summaries with the actual visual content to correct inaccuracies or misalignments. 
This automated yet refined workflow enables the production of high-quality, multi-granularity labels, ensuring both consistency and scalability across the dataset without the exhaustive cost of manual text writing.

\section{Experiments}

\subsection{Dataset Construction}
\label{sec:dataset_construction}
To facilitate a comprehensive evaluation of long-form video generation, we constructed a new, diverse, multi-source dataset. The foundational imagery is derived from the \textbf{COCO} dataset~\cite{lin2014microsoft}, which provides a rich repository of common objects and everyday scenes.

However, we identified significant gaps in COCO's scene coverage. To remedy this, we augmented the dataset with two supplementary sources: (1) \textbf{MIT Indoor Scenes}~\cite{quattoni2009recognizing}, to enrich the diversity of indoor environments; and (2) \textbf{Web-sourced Images}, a curated collection of high-resolution images depicting large-scale infrastructures, such as bridges and tunnels, which are notably scarce in existing benchmarks.

The complete dataset generation process (including pipeline-based video synthesis, described below) was computationally intensive, requiring approximately 20 days of processing time on a cluster of 32 NVIDIA A100 (80GB) GPUs. 
For the manual filtering stage, the entire process required approximately 2 person-days, with acceptance rates of 90\% for normal videos and 80\% for anomaly videos. 
This composite benchmark enables a more challenging and thorough assessment of a model's generalization capabilities and scene understanding.

\subsection{Implementation Details}
\label{sec:implementation_details}

Our proposed hierarchical generation pipeline (detailed in Section \ref{sec:pipeline}) employs several foundation models as its backbone:

\begin{itemize}
    \item \textbf{Video Generation Model:} For the core synthesis task, we employ the Wan~\cite{wan2025wanopenadvancedlargescale} video diffusion model as our T2V generator, valued for its high-fidelity output and temporal coherence.
    
    \item \textbf{Image-to-Storyline VLM:} We utilize Qwen2.5-VL-32B-Instruct~\cite{Qwen2.5-VL} as our primary VLM. It is responsible for parsing the initial frame and generating a structured, high-level "storyline" or sequence of events.
    
    \item \textbf{Event-to-Video-Level Model:} 
    To bridge the fine-grained to holistic semantic gap, Qwen3-8B~\cite{qwen3} acts as a semantic aggregator, synthesizing discrete event annotations into coherent video summaries. Subsequently, Qwen3-VL-32B~\cite{Qwen2.5-VL} serves as a VLM Refiner, cross-referencing these summaries with visual content to rectify any temporal or factual misalignments.

\end{itemize}
\begin{table*}[t]
  \centering
  \caption{Performance comparison on the Pistachio dataset for VAD. All methods are trained on the Pistachio training set. All metrics are reported in percentage (\%).}
  \label{tab:pistachio_results}
  
  \setlength{\tabcolsep}{4pt} 
  \renewcommand{\arraystretch}{1.0}
  \renewcommand{\theadfont}{\footnotesize\bfseries} 
  
  \resizebox{\textwidth}{!}{
  \begin{tabular}{l c c cc cc cc cc cc cc}
    \toprule
    \multirow{2}{*}{\textbf{Method}} & \multirow{2}{*}{\textbf{Year}} & \multirow{2}{*}{\textbf{Backbone}} & 
    \multicolumn{2}{c}{\makecell{\textbf{Public} \\ \textbf{Safety}}} & 
    \multicolumn{2}{c}{\makecell{\textbf{Accidents} \\ \textbf{Infrastructure}}} & 
    \multicolumn{2}{c}{\makecell{\textbf{Natural} \\ \textbf{Hazards}}} & 
    \multicolumn{2}{c}{\makecell{\textbf{Health} \\ \textbf{Medical}}} & 
    \multicolumn{2}{c}{\makecell{\textbf{Animal} \\ \textbf{Incidents}}} & 
    \multicolumn{2}{c}{\textbf{Overall}} \\
    
    \cmidrule(lr){4-5} \cmidrule(lr){6-7} \cmidrule(lr){8-9} \cmidrule(lr){10-11} \cmidrule(lr){12-13} \cmidrule(lr){14-15}
     & & & AUC & AP & AUC & AP & AUC & AP & AUC & AP & AUC & AP & AUC & AP \\
    \midrule
    
    RTFM \cite{rtfm}       & 2021 & I3D & 69.0 & 62.3 & 76.3 & 84.3 & \textbf{78.2} & \textbf{83.1} & 68.9 & 69.1 & 76.6 & \textbf{75.9} & 82.9 & 69.3 \\
    DR-DMU \cite{URDMU_zh} & 2023 & I3D & 66.5 & 63.4 & \textbf{79.8} & \textbf{88.4} & 75.1 & \textbf{83.1} & \textbf{76.0} & \textbf{78.8} & 69.8 & 64.0 & 81.5 & 71.5 \\
    MGFN \cite{chen2022mgfn}   & 2023 & I3D & 63.1 & 57.2 & 69.9 & 74.6 & 74.5 & 82.4 & 66.6 & 65.4 & 54.1 & 52.2 & 74.9 & 50.2 \\
    CLIP-TSA\cite{joo2023cliptsa} & 2023 & I3D & 66.9 & 58.5 & 78.7 & 83.3 & 76.4 & 82.9 & 69.9 & 62.1 & 64.5 & 56.5 & 76.9 & 55.2 \\
    CLIP-TSA\cite{joo2023cliptsa} & 2023 & ViT & 60.8 & 56.0 & 57.7 & 65.2 & 47.6 & 55.4 & 54.1 & 48.6 & 69.2 & 59.7 & 80.9 & 57.3 \\
    MULDE \cite{Micorek_2024_CVPR}  & 2024 & I3D & 50.1 & 40.1 & 62.0 & 58.4 & 52.0 & 49.2 & 57.5 & 37.8 & 55.1 & 43.9 & 63.4 & 34.9 \\
    VadCLIP\cite{wu2023vadclip}  & 2024 & ViT & 63.0 & 56.7 & 71.0 & 80.9 & 56.5 & 66.5 & 67.3 & 66.2 & 66.4 & 62.3 & 78.1 & 64.1 \\
    PEL4VAD\cite{pu2023learning}  & 2024 & ViT & \textbf{71.2} & \textbf{65.3} & 72.1 & 81.6 & 59.4 & 65.1 & 66.5 & 65.2 & \textbf{70.9} & 71.5 & \textbf{83.7} & 71.0 \\
    Fed-WSVAD\cite{wang2025federated} & 2025 & ViT & 62.2 & 54.8 & 79.7 & 86.1 & 74.6 & 83.0 & 71.2 & 64.6 & 49.3 & 47.3 & 83.2 & \textbf{71.9} \\
    VADTree\cite{li2025vadtreeexplainabletrainingfreevideo}  & 2025 & — & 64.99 & 44.21 & 63.37 & 42.15 & 53.73 & 62.76 & 68.09 & 52.74 & 57.30 & 39.00 & 72.52 & 25.15 \\
    
    \bottomrule
  \end{tabular}
  }
\end{table*}

\subsection{Analysis of Alternative Pipelines}
\label{sec:baseline_analysis}

To motivate our design, we first investigated several intuitive baseline pipelines for long-form video generation and identified their critical failure modes.

\begin{itemize}
    \item[\textbf{1.}] \textbf{Direct Long-Form Generation:} This one-shot approach tasks the model with generating the entire video from a single prompt. This method consistently suffers from two issues: (a) \textbf{Ghosting and Artifacts:} The model fails to maintain object identity and scene consistency over long durations. (b) \textbf{Prompt Forgetting:} The generated content progressively diverges from the initial prompt, indicating a severe semantic drift.

    \item[\textbf{2.}] \textbf{Fixed First-Last Frame (Looping):} This pipeline uses a first-last-frame-to-video model, setting the initial and final frames to be identical for seamless looping. This constraint, however, induces pathological motion artifacts. The model appears to ``rush" to satisfy the end-frame constraint, resulting in unnatural accelerations, "stuttering" movements, and physically implausible object trajectories as it attempts to force the scene back to its origin.

    \item[\textbf{3.}] \textbf{Naive Storyline Chaining:} This method generates video in short segments. For each new segment, the prompt is slightly modified based on the content of the previous segment's first frame. We found this approach lacks precise control. The model interprets the modified prompt as a simple "continuation" of the prior scene's state, rather than as an instruction to execute a new, distinct "event" as intended.
\end{itemize}

\subsection{Video Anomaly Detection}
\label{sec:vad_experiments}

We design three comprehensive sets of experiments to investigate the characteristics and utility of the Pistachio dataset. 
First, To evaluate the challenge our newly proposed Pistachio dataset poses to existing VAD methods, we selected several mainstream baseline methods for assessment.
Second, we conduct cross-dataset generalization experiments to examine the adaptability of models to new scenarios and perspectives. Moreover, to ensure our findings are not restricted by potential biases inherent to the Wan generation model, we incorporate additional evaluation sets generated by Hailuo and Cosmos. 
Finally, acknowledging that Pistachio features diverse viewpoints and cinematic styles—distinct from traditional fixed-angle surveillance datasets—we investigate the impact of data scaling and domain fusion by combining Pistachio with existing benchmarks to observe performance gains on real-world tasks.

\paragraph{Implementation Details.}

We selected nine widely used VAD methods as baselines: RTFM\cite{rtfm}, DR-DMU\cite{URDMU_zh}, MGFN\cite{chen2022mgfn}, MULDE\cite{Micorek_2024_CVPR}, CLIP-TSA\cite{joo2023cliptsa}, VadCLIP\cite{wu2023vadclip}, PEL4VAD\cite{pu2023learning}, Fed-WSVAD\cite{wang2025federated}, and VADTree\cite{li2025vadtreeexplainabletrainingfreevideo}. 
Following standard evaluation protocols, the training-based methods are trained on the Pistachio training set and evaluated on our test set, while the training-free method VADTree is directly deployed for evaluation without any parameter updates. 
Performance is measured using frame-level AUC and AP.
To prepare the data for the training-based methods, we extract frame-level features using pre-trained visual models, specifically I3D \cite{carreira2017quo} and ViT \cite{dosovitskiy2021imageworth16x16words}, following the processing approaches in \cite{rtfm} and \cite{wu2023vadclip}. 
For VADTree, we directly process the raw videos by leveraging pre-trained Generic Event Boundary Detection (GEBD) and video-language models following its original implementation.

\paragraph{Performance Comparison.}
As shown in Table~\ref{tab:pistachio_results}, we report the performance of baseline methods on the Pistachio test set.
Generally, most baseline methods exhibit suboptimal performance in the \textbf{Public Safety} and \textbf{Animal Incidents} categories. This performance gap suggests that current VAD models still struggle to capture fine-grained behavioral features and complex biological motion. Notably, while previous benchmarks often overlooked animal-related anomalies, the Pistachio dataset introduces these challenging scenarios, revealing the limitations of existing motion-based representations.

Among the evaluated methods, \textbf{PEL4VAD} \cite{pu2023learning}and our \textbf{Fed-WSVAD} \cite{wang2025federated} achieve superior results, which can be attributed to their distinct yet complementary strengths.
Commonly, both methods leverage vision-language association (e.g., CLIP-based features), allowing the models to integrate high-level semantic priors to disambiguate complex visual scenes. 
Individually, PEL4VAD excels by utilizing a prompt-enhanced learning mechanism to aggregate temporal context, which is particularly effective for categories requiring long-range dependency modeling. 
In contrast, our Fed-WSVAD benefits from a federated learning framework that captures a broader distribution of anomalies from decentralized clients. 
By aggregating diverse local knowledge into a global context-driven model, Fed-WSVAD achieves a more balanced and robust representation, leading to the highest Overall AP of 71.9\%.
The challenges posed by our dataset's enriched normal diversity are particularly evident when examining methods that lack semantic guidance.
The poor performance of MULDE (63.4\% AUC, 34.9\% AP) stems from its likelihood-based design, which struggles to define a tight boundary over highly variable normal data. More notably, the training-free VADTree yields a catastrophic AP drop to 25.15\% (72.52\% AUC); without in-domain training, its zero-shot vision-language reasoner is highly vulnerable to our enriched normal diversity, leading to severe false positives. 
These results underscore that neither traditional likelihood models nor zero-shot training-free approaches can robustly handle the complex normal variations and interactive anomalies introduced by Pistachio.
\begin{table*}[t]
\centering
\caption{Cross-dataset generalization performance across five target benchmarks and MSAD. Models are evaluated on both synthetic (Pistachio, Cosmos, Hailuo) and real-world (UCF-Crime, XD-Violence, MSAD) datasets. \textbf{Joint} denotes training on the union of all five datasets and is listed for reference. Best results among single-source training are in \textbf{bold}. AUC/AP metrics are reported in \%.}
\label{tab:wsvad_results_merged}
\setlength{\tabcolsep}{2pt} 
\renewcommand{\arraystretch}{1.0}
\resizebox{\textwidth}{!}{
\begin{tabular}{ll cc cc cc cc cc cc}
\toprule
\multirow{2}{*}{\textbf{Train Set}} & \multirow{2}{*}{\textbf{Method}} & \multicolumn{2}{c}{\textbf{Pistachio}} & \multicolumn{2}{c}{\textbf{Cosmos}} & \multicolumn{2}{c}{\textbf{Hailuo}} & \multicolumn{2}{c}{\textbf{UCF-Crime}} & \multicolumn{2}{c}{\textbf{XD-Violence}} & \multicolumn{2}{c}{\textbf{MSAD}} \\
\cmidrule(lr){3-4} \cmidrule(lr){5-6} \cmidrule(lr){7-8} \cmidrule(lr){9-10} \cmidrule(lr){11-12} \cmidrule(lr){13-14}
& & AUC & AP & AUC & AP & AUC & AP & AUC & AP & AUC & AP & AUC & AP \\
\midrule
\multirow{2}{*}{Pistachio}   & Fed-WSVAD & \textbf{83.23} & \textbf{71.86} & 81.09 & 78.76 & \textbf{83.97} & \textbf{79.46} & 55.69 & 10.14 & 70.11 & 44.51 & 81.68 & 24.28 \\
                             & CLIP-TSA  & 80.91 & 57.30 & 72.17 & 58.74 & 75.89 & 59.12 & 61.58 & 10.48 & 86.21 & 62.48 & 77.12 & \textbf{54.24} \\
\midrule
\multirow{2}{*}{UCF-Crime}   & Fed-WSVAD & 71.46 & 58.69 & 72.29 & 78.37 & 80.82 & 72.24 & \textbf{84.35} & \textbf{24.49} & — & — & — & — \\
                             & CLIP-TSA  & 61.06 & 39.80 & 71.54 & 58.66 & 68.71 & 70.84 & 82.56 & 21.62 & 81.93 & 51.75 & — & — \\
\midrule
\multirow{2}{*}{XD-Violence} & Fed-WSVAD & 71.61 & 62.14 & \textbf{81.67} & \textbf{81.43} & 83.36 & 71.64 & 78.65 & 19.64 & 92.50 & \textbf{77.04} & \textbf{86.44} & 32.76 \\
                             & CLIP-TSA  & 68.05 & 42.49 & 73.46 & 68.74 & 74.18 & 56.90 & 80.11 & 23.36 & \textbf{92.58} & 74.18 & 79.35 & 18.69 \\
\midrule
\midrule
Joint                        & CLIP-TSA  & 82.17 & 67.87 & 83.04 & 87.37 & 87.58 & 83.60 & 85.89 & 28.36 & 93.53 & 80.46 & 85.73 & 25.69 \\
\bottomrule
\end{tabular}
}
\end{table*}

\begin{table*}[t]
\centering
\caption{Impact of training data scaling on UCF-Crime test performance for CLIP-TSA. We evaluate performance by combining our Pistachio dataset with varying proportions of the UCF-Crime training set. All results are reported as AUC / AP (\%).}
\label{tab:data_scaling_ucf_reduced}
\setlength{\tabcolsep}{5pt} 
\renewcommand{\arraystretch}{1.0}
\scriptsize
\begin{tabular}{cc cc cc cc cc}
\toprule
\multicolumn{2}{c}{\textbf{Pistachio}} & \multicolumn{2}{c}{\textbf{UCF}} & \multicolumn{2}{c}{\textbf{Pist.+10\% UCF}} & \multicolumn{2}{c}{\textbf{Pist.+30\% UCF}} & \multicolumn{2}{c}{\textbf{Pist.+50\% UCF}} \\
\cmidrule(lr){1-2} \cmidrule(lr){3-4} \cmidrule(lr){5-6} \cmidrule(lr){7-8} \cmidrule(lr){9-10}
AUC & AP & AUC & AP & AUC & AP & AUC & AP & AUC & AP \\
\midrule
61.58 & 10.48 & 82.56 & 21.62 & 83.96 & 24.25 & 86.19 & 28.44 & \textbf{86.79} & \textbf{33.04} \\
\bottomrule
\end{tabular}
\end{table*}

\paragraph{Cross-Dataset Generalization and Bias Analysis}
As shown in Table \ref{tab:wsvad_results_merged}, the Pistachio dataset demonstrates strong cross-dataset generalization across diverse benchmarks, particularly on the synthetic Cosmos and Hailuo benchmarks as well as the real-world XD-Violence and MSAD datasets.
Notably, the competitive—and in some cases superior—performance of Pistachio-trained models compared to both UCF-Crime- and XD-Violence-trained models demonstrates promising transferability of our synthetic samples.
This outperformance can be attributed to Pistachio's greater scene diversity and balanced anomaly distribution, which together provide richer and more generalizable representations, underscoring its practical reliability as a training source for diverse real-world anomaly scenarios.

However, we observe that models trained solely on Pistachio show lower performance on UCF-Crime, due to the latter's low resolution and fixed-angle surveillance perspectives, as well as the inherent gap between synthetic and real-world data. 
To enhance portability to legacy surveillance scenarios, we explore a data scaling strategy. As illustrated in Table \ref{tab:data_scaling_ucf_reduced}, even with minimal fine-tuning using a small fraction of UCF-Crime data, the Pistachio-pretrained model achieves significant gains. Specifically, integrating Pistachio with just 10\% of the UCF training set outperforms a model trained on the full UCF dataset, demonstrating that Pistachio serves as a powerful foundational representation that complements traditional domain-specific data.
Finally, the joint training results in Table\ref{tab:wsvad_results_merged} further confirm the complementary nature of synthetic and real-world data: combining all datasets consistently achieves the strongest performance across all benchmarks, suggesting that Pistachio and real-world data are most powerful when used together.
However, we also acknowledge that Pistachio, as a synthetic dataset, carries inherent limitations in fully substituting real-world data, which we discuss in detail in Section~\ref{sec:limitations}.

Finally, to rule out generator-specific artifacts, we evaluate two dedicated VAD benchmarks built with Cosmos-2.5 and Hailuo-2.3. The Pistachio-trained models achieve consistent and significant improvements, raising the AUC from 71.54\% to 85.22\% on Cosmos and from 68.71\% to 82.34\% on Hailuo. These cross-generator enhancements confirm that Pistachio offers generalizable anomaly representations rather than overfitting to specific generator biases, establishing the robustness and diversity of our dataset.

\begin{table}[h]
\centering
\setlength{\abovecaptionskip}{0in}
\setlength{\belowcaptionskip}{0in}
\caption{Experiments of Pistachio datasets on Visual Anomaly Understanding, spanning multiple model families and scales. We differentiate understanding at the event-level and video-level outputs.}
\label{tab:vau}
\renewcommand{\arraystretch}{1.0}
\setlength{\tabcolsep}{2.3pt}
\scriptsize
\begin{tabular}{c|c|cc|c}
\toprule
\textbf{Model} & \textbf{Params} & \textbf{Event-level} $\uparrow$ & \textbf{Video-level} $\uparrow$ & \textbf{Avg.} $\uparrow$ \\
\midrule
InternVL3-1B        & 1B  & 28.25 & 23.04 & 25.65 \\
InternVL3-2B        & 2B  & 30.72 & 25.22 & 27.97 \\
Qwen2.5-VL-3B       & 3B  & 29.24 & 23.91 & 26.58 \\
LLaVA-Next-Video-7B & 7B  & 29.61 & 23.16 & 26.39 \\
Qwen2.5-VL-7B       & 7B  & 30.57 & 25.58 & 28.08 \\
Qwen3-VL-8B         & 8B  & 28.14 & \textbf{26.65} & 27.40 \\
InternVL3-8B        & 8B  & \textbf{32.56} & 25.65 & 29.11 \\
InternVL3-14B       & 14B & 32.29 & 26.49 & \textbf{29.39} \\
\bottomrule
\end{tabular}
\end{table}
\subsection{Video Anomaly Understanding}
To evaluate the benchmarks of video anomaly understanding, we adopt F1-Score as the main metric to measure the model's capability in comprehending anomalies across different temporal granularities. The F1-Score is calculated as the harmonic mean of precision and recall, providing a balanced assessment of the model's performance.

\noindent\textbf{Evaluation Protocol.} Following the hierarchical structure of our HIVAU-70k\cite{zhang2025holmes} dataset, we assess the model's understanding ability at two key levels: 1) \textbf{Event-level}, where the model analyzes anomaly events spanning multiple clips, requiring intermediate-term reasoning capabilities, and 2) \textbf{Video-level}, where the model comprehends the entire video narrative, demanding long-term contextual understanding. For each level, we measure the F1-Score by comparing generated descriptions with ground truth annotations, counting a prediction as correct if semantic similarity exceeds a predefined threshold.

\noindent\textbf{Comparison with State-of-the-Art Models.}As shown in Tab.~\ref{tab:vau}, we compare against several state-of-the-art Multimodal Large Language Models (MLLMs) spanning two families and a wide range of scales, including InternVL3 (1B--14B)\cite{chen2024internvl}, Qwen2.5-VL\cite{Qwen2.5-VL}, Qwen3-VL-8B\cite{qwen3technicalreport}, and LLaVA-Next-Video-7B\cite{zhang2024llavanextvideo}. The results reveal significant variations across model sizes and architectures, confirming that performance on Pistachio-VAU is sensitive to both model family and parameter scale. InternVL3-14B attains the best average score, while InternVL3-8B leads at the event level and Qwen3-VL-8B at the video level.

\noindent\textbf{Analysis.} The performance gap between event-level and video-level understanding reveals that models exhibit stronger capabilities in analyzing shorter temporal segments than comprehending extended video narratives. This observation supports our motivation for developing hierarchical instruction data, highlighting the challenge of maintaining contextual coherence over longer sequences. Furthermore, larger model parameters do not always guarantee better performance. For instance, the 8B-parameter Qwen3-VL achieves results comparable to 7B models, and within the InternVL3 family the gains from 8B to 14B are marginal, suggesting that model architecture and training data quality play crucial roles beyond raw parameter count.

\section{Scope and Limitations}
\label{sec:limitations}
The most fundamental limitation of Pistachio is the domain gap between synthetic and real-world data, which we characterise here openly. A model trained only on Pistachio transfers to real-world datasets with variable generalization performance, exposing a clear cross-domain discrepancy. On legacy, low-resolution surveillance data like UCF-Crime, CLIP-TSA reaches 61.58\% AUC / 10.48\% AP, dropping below the 82.56\% / 21.62\% obtained from training on UCF-Crime itself (Tables~\ref{tab:wsvad_results_merged} and~\ref{tab:data_scaling_ucf_reduced}). Although this performance variance is dataset-dependent—with Pistachio exhibiting relatively better transferability to MSAD and XD-Violence—the pronounced degradation observed on UCF-Crime demonstrates that models trained solely on our synthetic benchmark still encounter generalization limitations when facing specific real-world data distributions.

To help the community better navigate this gap, we frankly outline the conditions under which Pistachio can and cannot be expected to substitute for real-world data.
Specifically, Pistachio \textbf{cannot} serve as a direct substitute when the target task is exclusively restricted to traditional fixed-camera surveillance, or when it heavily relies on long-term character identity consistency. Despite our rigorous manual filtering, current video generation foundations still occasionally introduce subtle appearance shifts in characters over time. Conversely, Pistachio \textbf{can} effectively substitute for real-world data when the primary objective is to evaluate and enhance a model's capacity for detecting sudden, unexpected anomalies across generalized scenarios. 
For these tasks, Pistachio's rich anomaly categories, highly diverse storylines, and strictly balanced distribution provide a robust, bias-free evaluation bed that real-world benchmarks cannot replicate.

We accordingly offer concrete guidance for future users: Pistachio is best utilized as a scalable, high-diversity training foundation and a comprehensive gauge for generalized anomaly coverage. We recommend practitioners treat Pistachio-only metrics as a robust benchmark for broad anomaly comprehension, and perform modest in-domain fine-tuning when deploying to surveillance settings with low resolution or high-angle viewpoints.

\section{Conclusion}
In this paper, we introduce Pistachio, a new large-scale, Synthetic, Balanced, and Long-Form benchmark for VAD and VAU, accompanied by a highly automated and publicly released generation pipeline. Our comprehensive evaluation demonstrates that Pistachio's unique characteristics pose significant challenges to existing models, exposing their limitations. Furthermore, our synthetic approach provides precise ground truth for VAU, mitigating the heavy reliance on laborious manual annotation. We also provide concrete guidance for practitioners, recommending Pistachio as a scalable complement to real-world data and a foundation for fine-tuning rather than an unconditional substitute, and we hope it will serve as a valuable resource for future VAD and VAU research.

\section*{Acknowledgments}
This work was supported by the National Natural Science Foundation of China under Grant 62506235.

\bibliographystyle{splncs04}
\bibliography{main}

\clearpage
\appendix

\renewcommand{\thefigure}{A.\arabic{figure}}
\renewcommand{\thetable}{A.\arabic{table}}
\renewcommand{\theequation}{A.\arabic{equation}}
\setcounter{figure}{0}
\setcounter{table}{0}
\setcounter{equation}{0}

\begin{center}
  \vspace*{2em}
  {\Huge\bfseries Supplementary Material}
  \vspace*{2.5em}
\end{center}

\section{Details of the System Prompts.}
\label{sec:system}
The comprehensive system prompts (Tab \ref{tab:system_prompts_phase1}) are pivotal for precisely guiding the Large Language Model (LLM) through the multi-stage annotation workflow of the Pistachio dataset. These stage-specific prompts are designed to ensure structured and consistent output generation across four critical phases, aligning with the methodology described in the main paper:
\begin{figure*}[h!]
    \centering
    \includegraphics[width=\textwidth]{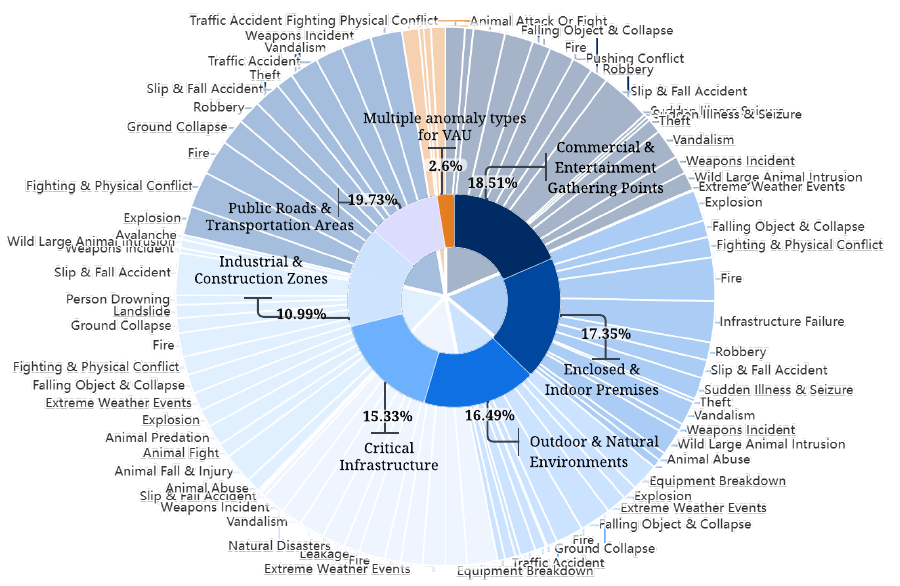}
    \caption{Distribution of all exception categories across all scenarios.}
    \label{fig:big_bing}
\end{figure*}

\begin{enumerate}
    \item \textbf{Scene-aware Classification} (Sec 3.2.1): A single, unified prompt template is used to categorize video content based on scene characteristics, which informs subsequent anomaly definition.
    
    \item \textbf{Anomaly Type Determination} (Sec 3.2.1): This stage employs six distinct system prompts, one dedicated to each of the six main scene categories. This granularity is essential for tailoring the anomaly definition process to the specific environmental and behavioral norms of each scene.
    
    \item \textbf{Multi-step Storyline Construction} (Sec 3.2.1): This phase utilizes a total of 68 prompt templates to cover the diverse scenarios in the dataset:
    \begin{itemize}
        \item Long-Form Anomaly Videos (31 templates): One template for each of the 31 fine-grained anomaly types to construct long-duration anomaly events.
        \item Short-Form Anomaly Videos (31 templates): One template for each of the 31 fine-grained anomaly types to construct short-duration anomaly events.
        \item Normal Videos (2 templates): Dedicated templates for generating storylines for both long-form and short-form normal (non-anomaly) videos.
        \item Multi-Anomaly Videos (4 templates): Templates for generating storylines containing multiple anomalies within a single video, covering the four specified complex anomaly categories.
    \end{itemize}
    We provide one representative template for each general type of scenario (e.g., one template for long-form anomaly videos, one for short-form normal videos) in Tab \ref{tab:system_prompts_phase2}.
    
\end{enumerate}

The full, verbatim text of all primary system prompt templates, including template structure and input variables, is comprehensively documented in Tab \ref{tab:system_prompts_phase1} and Tab \ref{tab:system_prompts_phase2}.

\begin{table}[h]
\centering
\caption{Comparison of video-level annotation quality between Pistachio-VAU and Holmes-VAU. Scores are presented as \textbf{Consistency / Richness} (1--10 scale), evaluated by various LVLMs.}
\label{tab:vau_comparison}
\small
\begin{tabularx}{\textwidth}{l @{\extracolsep{\fill}} c c}
\toprule
\textbf{Evaluator Model} & \textbf{Pistachio-VAU (Ours)} & \textbf{Holmes-VAU} \\
\midrule
Cosmos-Reason1-7B~\cite{nvidia2025cosmosreason1physicalcommonsense} & \textbf{7.43 / 7.81} & 6.43 / 5.99 \\
Qwen-2.5-7B~\cite{Qwen2.5-VL}       & \textbf{7.93 / 8.64} & 6.43 / 5.99 \\
Video-LLaVA-7B~\cite{lin2023video}    & 7.63 / \textbf{7.36} & \textbf{7.69} / 7.14 \\
\bottomrule
\end{tabularx}
\end{table}

\section{Dataset Characterization and Curation Details.}
\label{sec:dataset_analysis}

To further substantiate the quality, diversity, and robust generation methodology of the Pistachio dataset, we provide additional visualizations and analysis related to our curation process. Our complete generation pipeline, formalized in Algorithm~\ref{alg:pipeline}, demonstrates a systematic three-stage approach: Scene-Aware Classification, Anomaly Type Specification, and Multi-step Storyline Generation. This algorithmic framework ensures consistent and controllable video synthesis across all dataset samples.
Furthermore, Figure~\ref{fig:big_bing} provides a comprehensive, fine-grained view of the dataset's composition by mapping the distribution of all exception categories across the six primary scene scenarios, demonstrating the dataset's balanced and multi-faceted nature achieved through our structured pipeline.

\begin{table}[h]
\centering
\caption{VLM quality scores (1--5) across pipeline strategies. VQ: Visual Quality, TC: Temporal Coherence, AC: Anomaly Clarity, PA: Physical Authenticity.}
\label{tab:ablation}
\setlength{\tabcolsep}{4pt}
\renewcommand{\arraystretch}{1.0}
\scriptsize
\begin{tabular}{l cccc c}
\toprule
\textbf{Strategy} & \textbf{VQ} & \textbf{TC} & \textbf{AC} & \textbf{PA} & \textbf{Avg.} \\
\midrule
Direct Long-Form      & 4.0 & 3.0 & 2.0 & 2.0 & 2.75 \\
FLF                   & 3.9 & 4.3 & 4.2 & 3.8 & 4.04 \\
w/o Storyline         & 3.9 & 3.6 & 3.2 & 3.1 & 3.45 \\
\textbf{Ours}         & \textbf{4.0} & \textbf{4.5} & \textbf{4.4} & \textbf{3.8} & \textbf{4.17} \\
\bottomrule
\end{tabular}
\end{table}

\subsection{Pipeline Ablation Analysis}
\begin{figure*}[h!]
    \centering
    \includegraphics[width=\textwidth]{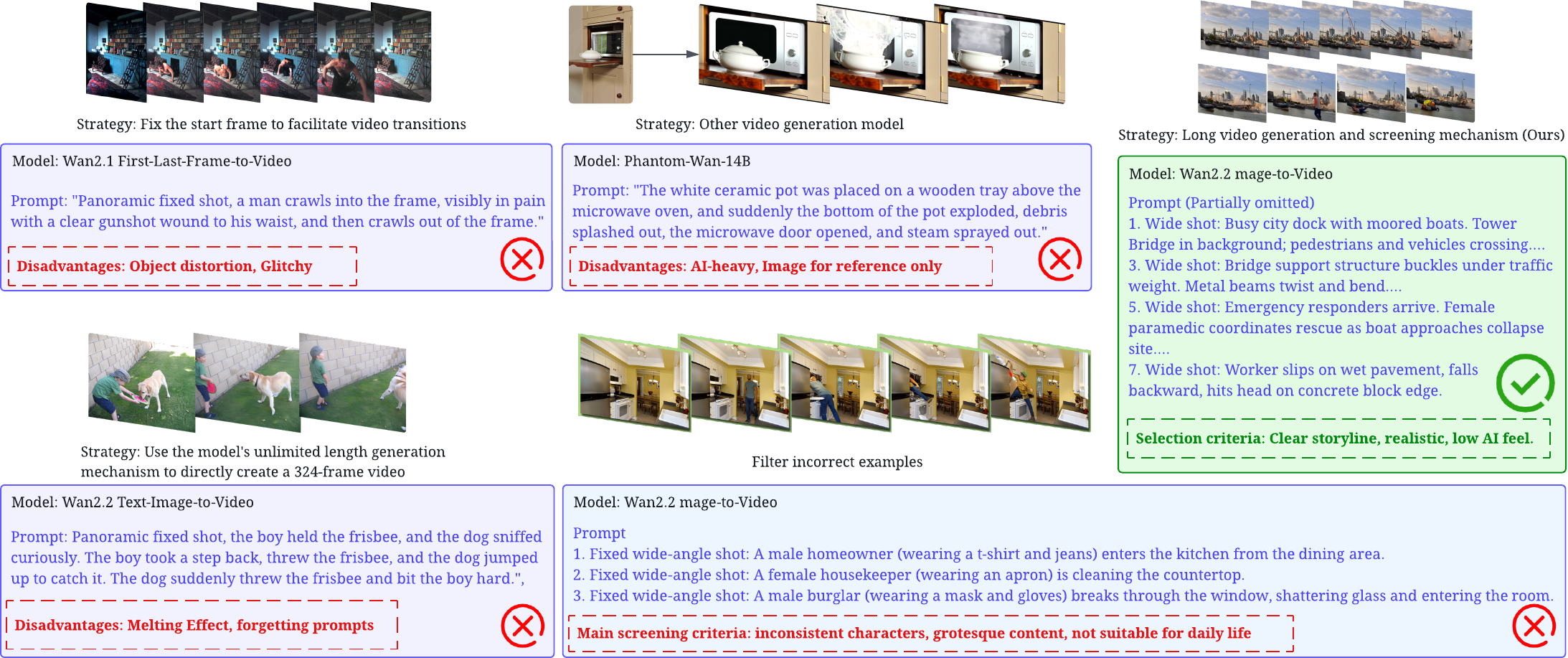}
    \caption{Comparison of Different Video Generation Schemes and Non-compliant Videos.}
    \label{fig:comparison}
\end{figure*}

Figure~\ref{fig:comparison} illustrates a detailed comparison of our proposed video generation scheme against alternative methods, specifically highlighting our effectiveness in handling and filtering non-compliant video outputs. The algorithm's scene-conditioned design (Lines 1-7) and hierarchical anomaly assignment strategy (Lines 8-18) enable precise control over the anomaly-scene pairing process, which is critical for generating realistic and contextually appropriate anomalies. To further quantify the quality advantage of our pipeline, Table~\ref{tab:ablation} reports VLM-based quality scores across four dimensions for each generation strategy, confirming that our full pipeline achieves the best overall performance.

\begin{table}[h]
\centering
\caption{Efficiency comparison across generation configurations.}
\label{tab:efficiency}
\setlength{\tabcolsep}{4pt}
\renewcommand{\arraystretch}{1.0}
\scriptsize
\begin{tabular}{l ccc}
\toprule
\textbf{Config} & \textbf{Time} & \textbf{VRAM} & \textbf{Speedup} \\
\midrule
Baseline (40-step)           & 349.7s & 14.3GB & 1$\times$ \\
4-step Distill               & 45.0s  & 10.2GB & 7.8$\times$ \\
4-step Distill + TeaCache    & 30.0s  & \textbf{3.2GB} & \textbf{11.7$\times$} \\
\bottomrule
\end{tabular}
\end{table}

\subsection{Generation Efficiency Analysis}
Table~\ref{tab:efficiency} reports the computational cost of our generation pipeline under different configurations. The baseline 40-step diffusion process requires 349.7s and 14.3GB VRAM per video. Adopting 4-step distillation reduces generation time by 7.8$\times$, and further combining it with TeaCache achieves an 11.7$\times$ speedup while reducing VRAM to only 3.2GB. We provide these results as a reference for practitioners seeking a more accessible alternative configuration for large-scale synthetic video generation.

\begin{figure}[H]
    \centering
    \includegraphics[width=0.6\linewidth]{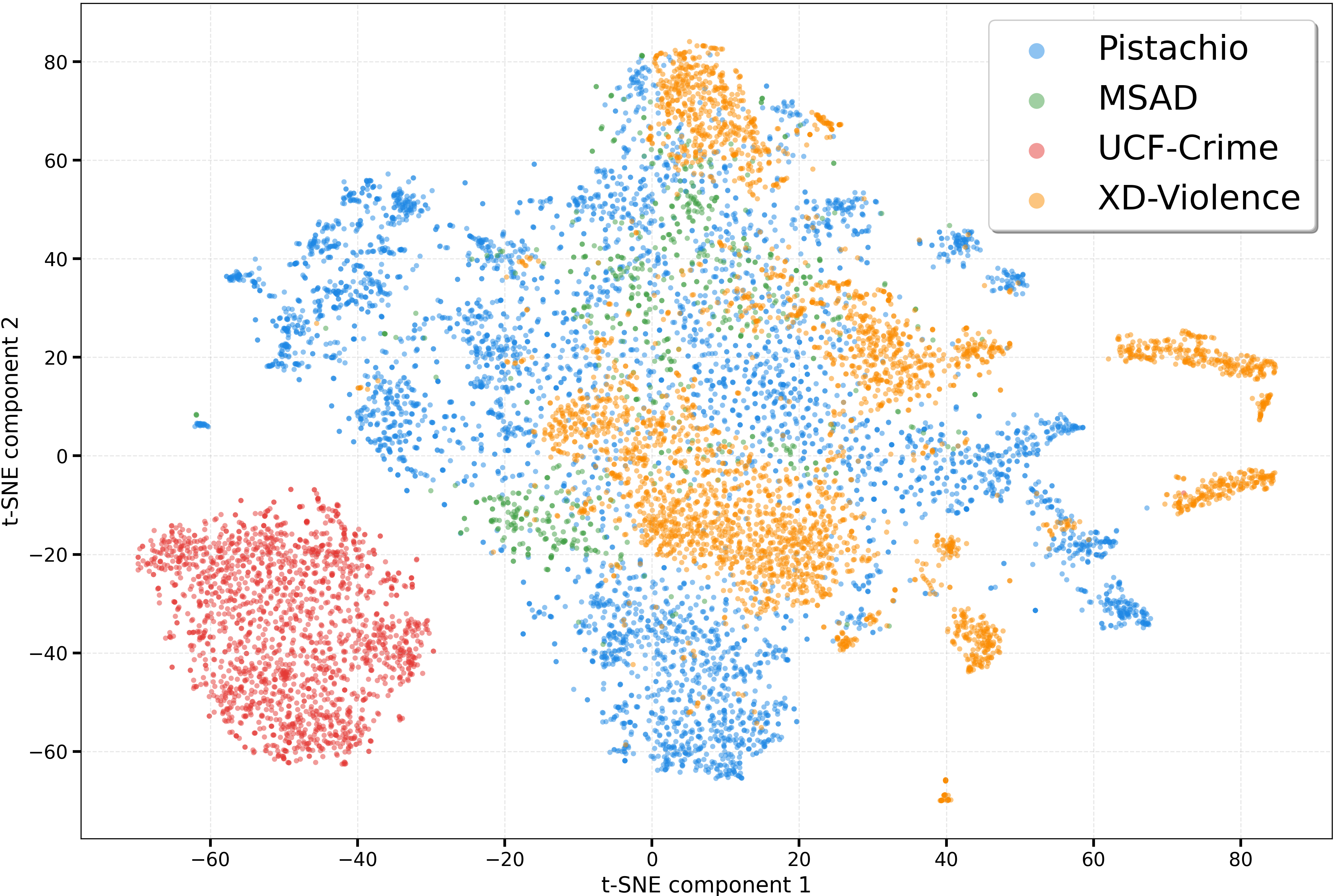}
    \caption{t-SNE Visualization of I3D Feature Distributions Across VAD Datasets.}
    \label{fig:tsne}
\end{figure}

\begin{table*}[h]
\centering
\caption{Cross-dataset generalization performance. Models are trained on a source dataset and tested on target datasets. Values in parentheses show the relative improvement when switching the training set from UCF-Crime to Pistachio.}
\label{tab:cross_dataset}

\setlength{\tabcolsep}{5pt} 
\renewcommand{\arraystretch}{1.0}
\scriptsize

\begin{tabular}{l cc cc cc}
\toprule
\multicolumn{7}{c}{\textbf{Training on UCF-Crime}} \\
\cmidrule(r){1-7}
\multirow{2}{*}{Method} & \multicolumn{2}{c}{UCF} & \multicolumn{2}{c}{UCF $\rightarrow$ XD} & \multicolumn{2}{c}{UCF $\rightarrow$ ShT} \\
\cmidrule(lr){2-3} \cmidrule(lr){4-5} \cmidrule(lr){6-7}
& AUC & AP & AUC & AP & AUC & AP \\
\midrule
VadCLIP  & 85.69 & 26.50 & 86.00 & 66.66 & 4.26 & 2.94 \\
CLIP-TSA & \underline{82.56} & \underline{21.62} & \underline{81.93} & \underline{51.75} & \underline{19.57} & \underline{3.27} \\
PEL4VAD  & \textbf{86.58} & \textbf{31.05} & \textbf{80.13} & \textbf{51.38} & \textbf{39.36} & \textbf{4.26} \\
\midrule
\midrule
\multicolumn{7}{c}{\textbf{Training on Pistachio (Ours)}} \\
\cmidrule(r){1-7}
\multirow{2}{*}{Method} & \multicolumn{2}{c}{Pistachio} & \multicolumn{2}{c}{Pist. $\rightarrow$ XD} & \multicolumn{2}{c}{Pist. $\rightarrow$ ShT} \\
\cmidrule(lr){2-3} \cmidrule(lr){4-5} \cmidrule(lr){6-7}
& AUC & AP & AUC & AP & AUC & AP \\
\midrule
VadCLIP  & 78.06 & 64.13 & 87.52 & 64.05 ($\downarrow$3.9) & 52.18 & 52.18 ($\uparrow$1675) \\
CLIP-TSA & \underline{80.91} & \underline{57.30} & \underline{86.21} & \underline{62.48} ($\uparrow$20.7) & \underline{69.15} & \underline{9.18} ($\uparrow$180.7) \\
PEL4VAD  & \textbf{83.70} & \textbf{70.96} & \textbf{85.94} & \textbf{63.42} ($\uparrow$23.4) & \textbf{78.43} & \textbf{23.04} ($\uparrow$440.8) \\
\bottomrule
\end{tabular}
\end{table*}

\subsection{Annotation Generation and Refinement}
A dedicated prompt template consolidates the event-level details produced in the preceding storyline-construction steps into the final structured, video-level annotations, strictly guiding the LLM to follow the required labeling format. A key property of this design is that it is \emph{causally inverted}: storylines are written before the video is generated and act as the generation prompt; we further refine these storylines via a vision-language model to produce the final ground-truth annotations, which makes annotation hallucination structurally impossible rather than something to be filtered out after the fact.
To validate annotation quality, we compare Pistachio-VAU against Holmes-VAU (using its UCF-based video data), as both rely on automated annotation pipelines. We evaluate two dimensions, Consistency and Richness, using multiple state-of-the-art models as evaluators. As shown in Table~\ref{tab:vau_comparison}, Pistachio-VAU attains higher scores on both metrics in the majority of cases, indicating that structured storyline guidance yields more coherent and informative annotations without manual intervention. We further conduct a human study on 100 videos with three annotators, obtaining consistency scores of 2.71 / 2.63 and inter-annotator agreement of $\kappa=0.76 / 0.72$ at the event / video level, confirming that the generated annotations are reliable under human assessment.

\subsection{Pistachio's Role in Generalization Evaluation}
As shown in Tab ~\ref{tab:cross_dataset}, the primary advantage of the Pistachio dataset is its effectiveness in \textbf{exposing the generalization gap} of existing VAD models when faced with new and diverse anomalies, which is crucial for assessing real-world applicability.
The reasons for this are threefold. First, our Pistachio dataset comprises 31 distinct anomaly types, over half of which are unique and not present in existing benchmarks like UCF-Crime. This diversity makes it an excellent testbed for evaluating out-of-distribution generalization capabilities. In contrast, UCF-Crime is primarily focused on crime-related activities, which limits its generalizability to datasets like ShT. Second, while UCF-Crime consists mostly of scenes with sparse crowds, our dataset encompasses a balanced variety of environments with both low and high pedestrian density. This composition enhances its generalization power towards crowded street scenarios, such as those found in ShT. For instance, the AUC of CLIP-TSA\cite{joo2023cliptsa} plummets from 82.56\% (on UCF) to 61.06\% (on Pistachio). This marked performance drop clearly indicates the model's failure to effectively recognize the semantic and visual characteristics of the novel, unseen anomalies unique to Pistachio.

This diversity gap is further corroborated by the t-SNE visualization of I3D features in Figure~\ref{fig:tsne}. Pistachio's feature distribution spans a substantially broader region of the embedding space, reflecting its rich scene diversity and balanced anomaly coverage. In contrast, UCF-Crime forms a compact, isolated cluster, consistent with its relatively low visual diversity stemming from fixed-angle, low-resolution surveillance footage with limited content variation. MSAD and XD-Violence occupy intermediate regions, partially overlapping with Pistachio, which explains their stronger cross-dataset transferability observed in Table~3 in the main paper.

The notable performance variations across the three methods on OOD test sets can be primarily attributed to their core design philosophy, particularly their utilization of Vision-Language Pre-training (VLP) knowledge.
VadCLIP \cite{wu2023vadclip} leverages a dual-branch structure that makes full use of the frozen CLIP model's fine-grained vision-language alignment. The visual features are enhanced by alignment with rich semantic language representations. This cross-modal knowledge acts as a strong semantic prior, providing a generalized understanding of "anomaly" that transcends dataset-specific visual features. Consequently, it achieves the best generalization performance to Pistachio's novel anomalies.
CLIP-TSA \cite{joo2023cliptsa} also utilizes the ViT-encoded visual features from CLIP, but it focuses on modeling temporal dependencies via a Temporal Self-Attention module. Crucially, it primarily relies on visual features and discards the rich semantic alignment information available through the language branch of CLIP. This visual-centric approach, while efficient, struggles to capture the novel semantic concepts in OOD anomalies, leading to a significant drop in its overall ranking ability (AUC) on datasets like Pistachio and ShanghaiTech.
PEL4VAD \cite{weak4} seeks to improve semantic discriminability through a Prompt-Enhanced Learning module, which integrates semantic priors using knowledge-based prompts extracted from an external knowledge base (ConceptNet). While this is a step beyond purely visual methods, its semantic priors are derived from a different source than CLIP's massive pre-training. Its generalization performance is expected to be an intermediate ground: better than purely visual models due to semantic enhancement, but potentially less robust than VadCLIP due to the difference in scale and breadth of the VLP knowledge utilized.

The results show that effective generalization to unseen anomalies in real-world scenarios (as simulated by the Pistachio dataset) requires leveraging generalized, semantically-rich cross-modal knowledge, rather than relying solely on visual features or simpler temporal modeling.

\begin{table}[h]
\centering
\caption{Detailed statistics of the Pistachio dataset (16 FPS). For the VAU task, the number of long videos is presented as \textit{Total (Single-Anomaly / Multi-Anomaly)}.}
\label{tab:dataset_stats_compact}
\setlength{\tabcolsep}{6pt}
\renewcommand{\arraystretch}{1.0}
\scriptsize

\begin{tabular}{l ccc c}
\toprule
\textbf{Pistachio (16FPS)} & \multicolumn{3}{c}{\textbf{VAD Task}} & \textbf{VAU Task} \\
\cmidrule(r){2-4} \cmidrule(l){5-5} 
& \textbf{Train} & \textbf{Test} & \textbf{Total} & \textbf{Total} \\
\midrule
\# Anomaly Frames & 420,705 & 72,546 & 493,251 & -- \\
\# Normal Frames & 1,123,841 & 59,730 & 1,183,571 & -- \\
\midrule
\textbf{\# Total Frames} & \textbf{1,544,546} & \textbf{132,276} & \textbf{1,676,822} & \textbf{517,514} \\
\midrule
\# Scenes & 3,750 & 402 & 3,896 & 1,277 \\
\# Main Anomaly Types & 31 & 30 & 31 & 31 \\
\midrule
\# Long Videos & 1,158 & 106 & 1,264 & 523 (487/35) \\ 
\# Short Videos & 3,402 & 296 & 3,698 & 862 \\
\bottomrule
\end{tabular}
\end{table}

\begin{table*}[t]
\centering
\caption{System prompts for Scene-aware Classification, Anomaly Type Determination, and Event-to-Video-level Annotation Generation phases of the Pistachio dataset generation pipeline.}
\label{tab:system_prompts_phase1}
\setlength{\tabcolsep}{5pt} 
\renewcommand{\arraystretch}{1.0}
\scriptsize
\begin{tabularx}{\textwidth}{l X}
\toprule
\textbf{Prompt Type} & \textbf{Prompt Content} \\
\midrule
\multirow{12}{3cm}{\textbf{Scene-aware Classification}} & 
You are a professional image scene analysis expert. Your sole task is to accurately classify the images provided by users into the following six fixed scene categories. You must strictly adhere to the rules: \\
& \textbf{No creating new categories:} You may only choose one of the following six options and must not invent new category names. \\
& \textbf{No explanations:} Your response must not include any analysis process, reasoning, possibilities, or confidence levels. \\
& \textbf{No multi-labeling:} Each image has one and only one most matching scene. \\
& \textbf{Response format:} You must and can only return a pure, unmodified category name. \\
& \textbf{The six scene categories:} \\
& 1. Public Roads \& Transportation Areas \\
& 2. Enclosed \& Indoor Premises \\
& 3. Commercial \& Entertainment Gathering Points \\
& 4. Industrial \& Construction Zones \\
& 5. Outdoor \& Natural Environments \\
& 6. Critical Infrastructure \\
\midrule
\multirow{8}{3cm}{\textbf{Anomaly Type Determination} \\ (6 prompts, one per scene)} & 
\textbf{Example for Commercial \& Entertainment Gathering Points:} \\
& You are an expert in multi-image analysis and event inference for "Commercial \& Entertainment Gathering Points." You will receive a set of image files. Your task is to assign the most likely and specific anomalous events that will happen in the future for each image according to the following priorities: first, identify the most probable anomalous events. Second, ensure that the types of events are relatively evenly distributed across the entire image set. \\
& \textbf{Abnormal Event Categories (15 specific events):} Theft, Fighting \& Physical Conflict, Weapons Incident, Robbery, Animal abuse, Slip \& Fall Accident, Pushing Conflict, Sudden Illness \& Seizure, Fire, Vandalism, Explosion, Falling Object \& Collapse, Animal Attack or Fight, Wild Large Animal Intrusion, Other. \\
& \textbf{Output Requirements:} Output must be a strict numbered list. The [event\_name] must be a single, lowercase string with spaces replaced by underscores. \\
& \textit{Note: Similar prompts exist for the other 5 scene categories with category-specific anomaly types.} \\
\midrule
\multirow{3}{3cm}{\textbf{Event-level to Video-level Annotation}} & 
Summarize the following series of events into a single, concise video-level prompt. The summary should combine the main actions, characters, and settings into a continuous narrative. Please output only the summary content without any prefix or explanation. \\
& \textbf{Input:} Event sequence from Multi-step Storyline Construction \\
& \textbf{Output:} Consolidated video-level description \\
\bottomrule
\end{tabularx}
\end{table*}

\begin{table*}[t]
\centering
\caption{Representative system prompt templates for the \textbf{Multi-step Storyline Construction} phase of the Pistachio dataset generation pipeline. These templates guide the LLM to construct coherent narratives for different video types.}
\label{tab:system_prompts_phase2}
\setlength{\tabcolsep}{5pt} 
\renewcommand{\arraystretch}{1.0}
\scriptsize
\begin{tabularx}{\textwidth}{l X}
\toprule
\textbf{Prompt Type} & \textbf{Prompt Content} \\
\midrule
\multirow{5}{3cm}{\textbf{Normal Storyline} \\ \textbf{(Short-Form)}} 
& You are an expert in crafting video description prompts. Your task is to transform the image content provided by the user into multiple coherent dynamic segments... Specific requirements are as follows: \\
& \textit{[Key requirements: 3-6 second segments; 2-3 paragraphs; "Fixed wide shot"; continuous, normal dynamic scenes;} \\
& \textit{total output $\leq$ 100 words.]} \\
& Example 1: 1. Fixed wide shot. A young woman carrying a stack of books walks out from a library entrance... \\
\midrule
\multirow{6}{3cm}{\textbf{Normal Storyline} \\ \textbf{(Long-Form)}} 
& You are an expert in creating prompts for segmented video stories. Your task is to construct a coherent long narrative comprising 7-8 independent video segments based on the image content... Specific requirements are as follows: \\
& \textit{[Key requirements: 4-6 second segments; 7-8 paragraphs; "Fixed wide-angle shot:"; focus on ordinary daily scenes;} \\
& \textit{specify gender/description for all persons; total output $\leq$ 150 words.]} \\
& Example 1: 1. Fixed wide-angle shot: Elderly man reads newspaper on park bench, young mother pushes stroller entering from right... \\
\midrule
\multirow{5}{3cm}{\textbf{Anomaly Storyline} \\ \textbf{(Short-Form)}}
& You are an expert in crafting video description prompts. Your task is to take an image input by the user and use reasonable imagination to bring the image to life, emphasizing potential dynamic anomalies... Specific requirements are as follows: \\
& \textit{[Key requirements: 5 second segments; 2-3 segments; "panoramic fixed shot"; results must revolve around specific anomalies like "collision",} \\
& \textit{avoiding vague words; total output $\leq$ 100 words.]} \\
& Example 1: 1.Panoramic fixed shot: A blue sedan drives through an intersection when suddenly a red truck runs the stoplight, colliding with the sedan's passenger side... \\
\midrule
\multirow{5}{3cm}{\textbf{Anomaly Storyline} \\ \textbf{(Long-Form)}} 
& You are an expert in creating prompts for segmented video stories. You will receive a specific abnormal event type requirement from the user: "Infrastructure Failure"... Based on this, you need to construct a storyline distributed across 7-8 independent video segments... \\
& \textit{[Key requirements: 7-8 segments; "Fixed wide-angle shot:"; failure must be clearly visible and show public impact;} \\
& \textit{first segment must establish normal operation; avoid words like "almost".]} \\
& Example 1: 1. Fixed wide-angle shot: A busy intersection with traffic lights functioning and vehicles (mixed types) flowing normally... \\
\midrule
\multirow{6}{3cm}{\textbf{Multi-Anomaly Storyline}} 
& You are an expert in creating prompts for segmented video stories. You will receive a specific requirement: a storyline containing TWO sequential abnormal events - "Explosion" followed by "Falling Object \& Collapse"... Based on this, you need to construct a storyline distributed across 6-8 independent video segments...\\
& \textit{[Key requirements: 6-8 segments; "Fixed wide-angle shot:"; must depict TWO sequential abnormal events;} \\
& \textit{first segment must be normal; must show clear blast effects and subsequent collapse/falling objects.]} \\
& Example 1: 1. Fixed wide-angle shot: A factory building with workers (three males) operating machinery near storage racks... \\
\bottomrule
\end{tabularx}
\end{table*}

\subsection{Discussion and Future Directions}
The comparative analysis across both in-distribution Tab 2 and cross-dataset generalization Tab ~\ref{tab:cross_dataset} experiments reveals several critical directions for advancing Video Anomaly Detection systems. First, the substantial performance gap between training and testing on Pistachio versus traditional datasets like UCF-Crime underscores the urgent need for models that can handle diverse, balanced anomaly distributions beyond crime-centric scenarios. The success of VadCLIP in cross-dataset generalization, attributed to its exploitation of frozen CLIP's vision-language alignment, suggests that future research should prioritize the integration of large-scale vision-language pre-training knowledge to capture generalizable semantic concepts of "anomaly" rather than dataset-specific visual patterns. Meanwhile, the relative weakness of methods like MULDE on Pistachio's high-variability normal samples indicates that likelihood-based approaches struggle when normal data exhibits complex, diverse behaviors, pointing toward the necessity of developing methods that can distinguish subtle behavioral deviations rather than relying solely on large visual changes.
Looking forward, the dramatic improvements in cross-dataset performance when training on Pistachio (e.g., PEL4VAD\cite{weak4} achieving 440.8\% AP improvement on ShanghaiTech compared to UCF-Crime training) demonstrate that synthetic, balanced datasets can serve as effective pre-training resources for real-world deployment. Future VAD architectures should embrace multi-modal learning paradigms that combine rich textual semantic priors with temporal visual modeling, while also incorporating mechanisms for handling long-form videos with complex, multi-event narratives. The Pistachio benchmark's introduction of novel anomaly categories like landslides, animal predation, and infrastructure failures—absent from existing datasets—highlights the need for open-vocabulary detection capabilities that can recognize previously unseen anomaly types through semantic reasoning rather than purely visual matching. To successfully transition models from synthetic data to real-world applications, future research should combine the advantages of generation-based datasets (scalability, balance, and controllability) with domain adaptation strategies that can bridge the visual and contextual differences between synthetic and authentic surveillance footage. Moreover, we plan to extend our generation pipeline to enable precise frame-level control over anomaly onset and termination timestamps, addressing the fundamental limitations of manual annotation where human labelers often disagree on the exact boundaries of anomalous events. This capability will provide objectively defined, pixel-accurate temporal ground truth that eliminates annotator subjectivity and enables more reliable evaluation of frame-level detection algorithms.

\begin{algorithm}[t]
\caption{Scene-Conditioned Anomaly Storyline Generation}
\label{alg:pipeline}

\textbf{Input:} Image set $\mathcal{I} = \{I_1, \cdots, I_N\}$, Vision-Language Model $\mathcal{M}(\cdot)$, Scene configuration $\mathcal{C}_{scene}$

\textbf{Output:} Image-storyline pairs $\{(I_i, \mathcal{S}_i)\}_{i=1}^N$

\begin{algorithmic}[1]

\State \textbf{Stage 1: Scene Classification}
\For{$i = 1$ to $N$}
    \State Extract scene categories $C = \{c_1, \cdots, c_K\} \cup \{\text{``Other''}\}$ from $\mathcal{C}_{scene}$
    \State $\hat{c}_i \gets \mathcal{M}(I_i, \text{prompt}_{\text{scene}})$ using Eq. 1
    \State Assign $I_i \to \mathcal{G}_{\hat{c}_i}$
\EndFor
\State \textbf{Stage 2: Anomaly Type Assignment}
\State Initialize anomaly mapping $\mathcal{A}: I_i \mapsto a_i$
\For{each scene category $c_j \in C$}
    \If{$|\mathcal{G}_{c_j}| > 0$}
        \State $A_j \gets \{a_{j,1}, \cdots, a_{j,M_j}\}$ from $\mathcal{C}_{scene}[c_j]$
        \State $\phi_{j} \gets \text{BuildPrompt}(c_j, A_j)$ using Eq. 2
        \For{each $I_i \in \mathcal{G}_{c_j}$}
            \State $\hat{a}_i \gets \mathcal{M}(I_i, \phi_j)$
            \State $\mathcal{A}[I_i] \gets \hat{a}_i$
        \EndFor
    \EndIf
\EndFor
\State \textbf{Stage 3: Storyline Generation}
\For{$i = 1$ to $N$}
    \State Retrieve $c_i \gets \text{GetScene}(I_i, \mathcal{G})$ and $a_i \gets \mathcal{A}[I_i]$
    \State $p_i \gets \mathcal{C}_{scene}[c_i][a_i]$
    \State $\psi_i \gets \text{FormatPrompt}(p_i, I_i)$ using Eq. 3
    \State $\mathcal{S}_i \gets \mathcal{M}(I_i, \psi_i)$ using Eq. 4
\EndFor
\State \Return $\{(I_i, \mathcal{S}_i)\}_{i=1}^N$

\end{algorithmic}
\end{algorithm}

\end{document}